\documentclass[conference]{IEEEtran}
\IEEEoverridecommandlockouts
\usepackage{cite}
\usepackage{amsmath,amssymb,amsfonts}
\usepackage{algorithmic}
\usepackage{graphicx}
\usepackage{textcomp}
\usepackage{subfigure}
\usepackage{babel}
\usepackage{xcolor}
\usepackage{xurl}
\usepackage{multicol}
\def\BibTeX{{\rm B\kern-.05em{\sc i\kern-.025em b}\kern-.08em
    T\kern-.1667em\lower.7ex\hbox{E}\kern-.125emX}}
\begin{document}

\title{Extending Machine Learning-Based Early Sepsis Detection to
Different Demographics\\
}


\author{\IEEEauthorblockN{Surajsinh Parmar}
\IEEEauthorblockA{\textit{SpassMed Inc.} \\
Toronto, Canada \\
suraj.parmar@spassmed.ca}
\and
\IEEEauthorblockN{Tao Shan}
\IEEEauthorblockA{\textit{University of Waterloo} \\
Waterloo, Canada \\
t4shan@uwaterloo.ca}
\and
\IEEEauthorblockN{San Lee}
\IEEEauthorblockA{\textit{SpassMed Inc.} \\
Toronto, Canada \\
sanlee@spassmed.ca}
\and
\IEEEauthorblockN{Yonghwan Kim}
\IEEEauthorblockA{\textit{Spass Inc.} \\
Seoul, Korea \\
kyh@spass.ai}
\and
\IEEEauthorblockN{Jang Yong Kim}
\IEEEauthorblockA{\textit{St. Mary's Hospital} \\
Seoul, Korea \\
vasculakim@catholic.ac.kr}
}

\maketitle

\begin{abstract}
Sepsis requires urgent diagnosis, but research is predominantly focused on Western datasets. In this study, we perform a comparative analysis of two ensemble learning methods, LightGBM \cite{NIPS2017_6449f44a} and XGBoost \cite{10.1145/2939672.2939785}, using the public eICU-CRD dataset and a private South Korean St. Mary’s Hospital's dataset. Our analysis reveals the effectiveness of these methods in addressing healthcare data imbalance and enhancing sepsis detection. Specifically, LightGBM shows a slight edge in computational efficiency and scalability. The study paves the way for the broader application of machine learning in critical care, thereby expanding the reach of predictive analytics in healthcare globally.
\end{abstract}

\begin{IEEEkeywords}
Machine Learning, Spesis Detection, Classification
\end{IEEEkeywords}

\section{Introduction}
Sepsis demands rapid and precise diagnosis but faces challenges like general symptoms and lack of clear biomarkers \cite{nelson2014biomarkers}, \cite{celik2022diagnosis}. Real-time vital signs can offer clues for early detection. Our study focuses on using machine learning to identify sepsis risk using vital signs. We also explore if demographic factors like age, gender, and ethnicity could improve model generalizability. 

The challenge of transfer learning across different healthcare settings is also considered. We test our approaches on two datasets: the eICU-CRD \cite{pollard2018eicu} and a private dataset from the St. Mary's hospital South Korea. Both LightGBM and XGBoost show strong performance, especially in handling data imbalance, confirming their potential in healthcare analytics.

\section{Related Works}
In recent years, various machine learning, time series forecasting, and deep learning techniques have been applied to sepsis prediction and detection, often utilizing MIMIC-III and eICU-CRD datasets for model development and validation \cite{johnson2016mimic}, \cite{pollard2018eicu}. 

Studies like those by \cite{10.3389/fmed.2021.662340} and \cite{hou2020predicting} have specifically harnessed the power of XGBoost for outcome prediction in ICU settings, including 30-day mortality rates among sepsis patients. These works have consistently outperformed traditional models such as logistic regression and SAPS-II, thereby highlighting the transformative potential of machine learning in clinical care.

Traditional approaches to sepsis prediction have often employed logistic regression, SVM, and decision trees as foundational methodologies \cite{7_bloch2019machine}, \cite{8_gultepe2014from}. The potential of ensemble methods in early sepsis detection has also been realized, as demonstrated by the use of gradient boosted trees and random forests in recent studies \cite{9_BARTON201979}, \cite{3_FORKAN2017244}. Reference \cite{bhatti2023interpreting} explored the forecasting of vital signs using deep learning models and  \cite{salimiparsa2023investigating} experimented with interpretibility of classifiers focusing on explainable model for predictions.

Emerging trends include the adoption of Hierarchical Temporal Memory for the analysis of vital signs \cite{2_staffs6212}, and the use of Transformers for time series data prediction \cite{10_zeng2022transformers}. Researchers have also given attention to the quality of data used for these predictions. Techniques for outlier detection in electronic health records have been introduced \cite{1_estiri2019clustering}, and the challenges presented by sensor faults have been addressed \cite{4_6655254}.

Complementary to model-based techniques, data processing strategies aimed at enhancing predictive accuracy have been developed, such as physiologic reasoning for shock state diagnosis \cite{5_lighthall2011physiologic} and multi-site studies on early warning systems for sepsis \cite{6_adams2022prospective}.

\section{Data Preprocessing}

Our study leverages data from the eICU-CRD dataset \cite{pollard2018eicu}, supplemented with the St. Mary's Hospital's records for external validation. The combined dataset consists of 6,334 patients diagnosed with sepsis, aggregated into 17,225 non-overlapping units, each unit representing a 6-hour window of vital sign data, totaling 1,240,200 timestamps. 

We uniformized the timestamps to 5-minute intervals across all vital signs, which include systolic blood pressure (systolicbp), diastolic blood pressure (diastolicbp), mean blood pressure (meanbp), heart rate (heartrate), respiration rate (respiration), peripheral oxygen saturation (spo2), pulse pressure (pp), and the Glasgow Coma Scale score (gcs). Missing values were managed through forward-filling and backward-filling techniques. We performed data sanity checks, and units with implausible vital sign values were eliminated. This left us with 10,743 valid and non-overlapping segments, and our prediction target is the likelihood of sepsis onset within the 3 hours succeeding each 6-hour window.

It's worth noting that 24.4\% of these units were labeled as sepsis-positive. This imbalance necessitates tailored modeling approaches to account for the skewed class distribution, a crucial detail for deep learning practitioners.

\subsection{Feature Engineering}

Given the complexity and the high-dimensionality of time-series medical data, an array of feature engineering techniques was adopted. First, we incorporated time lags into the data to capture temporal patterns and trends. Statistical metrics such as mean, standard deviation, maximum, minimum, kurtosis, median, and skewness were computed for each vital sign variable across varying window periods to encapsulate the data's distributional properties and temporal variations. 

Fourier Transform techniques were also applied to identify underlying cyclical patterns in the vital signs. This was particularly useful for capturing periodic behaviors like circadian rhythms that are less obvious in raw data but can be critical for prediction in a medical context.

Another interesting aspect was the computation of lagged differences to identify the rate of change in the vital signs over time, capturing both the magnitude and direction of changes, which are particularly crucial for early sepsis identification.

The feature matrix post-engineering had $10,743 \textit{ groups} \times 138 \textit{ features}$, incorporating the computed lags and statistical attributes.

\subsection{Implementation Details}

Evaluating a model on time-series data presents unique challenges, especially in preventing data leakage. Our evaluation strategy employed StratifiedGroupKFold for splitting the data, maintaining the separation of groups (patients) and preserving the class distribution across training and validation sets.

To prevent leakage during feature generation, we ensured that features for a given timestamp were solely derived from data prior to that timestamp. This is a critical consideration when working with temporal data, as ignoring it could introduce look-ahead bias, thereby inflating performance metrics.

Our framework ensures a rigorous evaluation process, aimed at providing an honest assessment of our model's capability to predict sepsis onset based on the available vital signs and demographic data. This meticulous approach to implementation is geared towards deep learning practitioners who understand the nuances of working with imbalanced and temporally sensitive data.

\section{Methodology - Modeling}
For sepsis prediction, we employ tree-based models with a focus on XGBoost and LightGBM, owing to their balance of interpretability and computational efficiency. Optuna \cite{akiba2019optuna} serves as the hyperparameter tuning framework, facilitating the fine-tuning of crucial parameters in both models. Both models employ Gradient Boosted Decision Trees and Random Forests as part of their tree-building strategies.

\subsection{Evaluation Metrics}
Given that only 24.4\% of the dataset represents sepsis-positive cases, standard accuracy metrics are not sufficient. We adopt the area under receiver operating characteristic curve (AUC-ROC) as the primary metric due to its efficacy in handling class imbalances. Additionally, precision, recall, and F1 scores are considered to provide a well-rounded evaluation. A high recall is prioritized to minimize the risk of missing any true sepsis cases, which is clinically crucial.

\subsection{Models}
XGBoost operates through iterative tree boosting and effectively manages missing values and class imbalances, making it well-suited for complex, tabular data. On the other hand, LightGBM uses a unique ``leaf-wise" tree growth strategy, allowing for more complex models that are especially effective for large-scale datasets. Both models are well-aligned with the challenges posed by our dataset in predicting sepsis onset. For additional context and comparative analysis, Support Vector Machine (SVM), Random Forest, Logistic Regression, and Long Short-Term Memory (LSTM) models were also explored.

\section{Results}

Optuna was employed for hyperparameter tuning of LightGBM and XGBoost models. This process enhanced their predictive capability for sepsis onset. LightGBM offered advantages in treating missing data, while XGBoost reduced model overfitting.

\subsection{Model Performance on eICU Dataset}

We evaluated multiple machine learning algorithms on the eICU dataset using key metrics: AUC-ROC, Precision, Recall, and F-1 Score. LightGBM and XGBoost emerged as top performers in AUC-ROC, thus showing a superior ability to distinguish between sepsis and non-sepsis cases. Random Forest had the highest Precision, but LightGBM yielded the best F-1 Score, a balanced measure between precision and recall. While LSTM showed high scores across metrics, its computational expense was considerably higher.

\begin{table}[htbp]
\centering
\scriptsize
\caption{Performance comparison for eICU dataset on out of sample 5 fold cross validation}
\begin{tabular}{cccccc}
\hline
\textbf{Model} & \textbf{AUC} & \textbf{Precision} & \textbf{Recall} & \textbf{F-1 Score} & \textbf{Time(s)}\\
\hline
\textbf{LightGBM} & 0.964 & 0.839 & 0.827 & 0.832 & 0.167 \\
\textbf{XGBoost} & 0.965 & 0.832 & 0.840 & 0.836 & 1.196\\
SVM & 0.901 & 0.788 & 0.593 & 0.676 & 2.400\\
Random Forest & 0.948 & 0.881 & 0.731 & 0.806 & 4.817\\
Logistic Regression & 0.902 & 0.786 & 0.601 &  0.681 & 0.188 \\
LSTM & 0.967 & 0.869 & 0.821 & 0.844 & 376.850 \\

\hline
\end{tabular}
\label{tab:performance_comparison_1}
\end{table}

Overall, the comparison suggests that LightGBM and XGBoost models has great AUC-ROC and efficiency to choose. There are trade offs between Precision and Recall when we changing model parameters and prediction thresholds. LSTM has the best performance but it has high computational cost. Besides, GBDT models have shown to be robust with tabular data and missing values.

\subsection{Additional Testing on Hospital's Datasets}

To ensure broader applicability, models were also tested on St. Mary's Hospital datasets. Both eICU and Hospital datasets share similar types of data—Vital Signs and Demographics—but differ in data capture frequency, sample size, and imbalance in labels.

\begin{table}[htbp]
\centering
\scriptsize
\caption{Performance on the datasets - Light GBM}
\begin{tabular}{cccccc}
\hline
\textbf{Model} & \textbf{AUC} & \textbf{Precision} & \textbf{Recall} & \textbf{F-1 Score} & \textbf{Modeltype} \\
\hline
eICU & 0.964 & 0.839 & 0.827 & 0.832 & LightGBM\\
eICU & 0.965 & 0.832 & 0.840 & 0.836 & XGBoost\\
Hospital & 0.677 & 0.617  & 0.677 &  0.638 & LightGBM\\
Hospital & 0.633 & 0.727  &0.633& 0.666 & XGBoost\\
Hospital + eICU & 0.776 & 0.819 & 0.776 & 0.796 & LightGBM\\
Hospital + eICU & 0.767 & 0.736 & 0.767 & 0.750 & XGBoost\\
\hline
\end{tabular}
\label{tab:performance_comparison_4}
\end{table}

\subsection{AUC-ROC Curve Interpretation}

The AUC-ROC curves were plotted to visualize how well the model performs when the costs of false positives and false negatives are considered. This helps in understanding the trade-off between the True Positive Rate (also known as Recall) and the False Positive Rate.

\begin{figure}[!h]
    \centering
    \subfigure[XGBoost results]
    {
        \includegraphics[width=0.8\columnwidth]{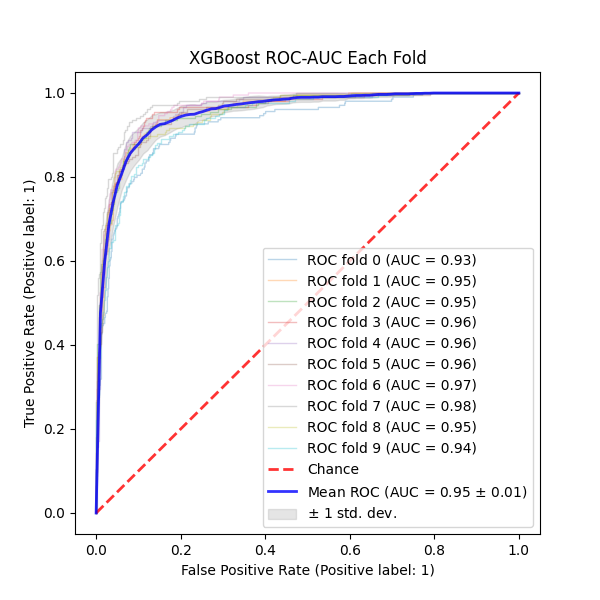}
        \label{fig:first_sub}
    }
    \subfigure[LightGBM results]
    {
        \includegraphics[width=0.8\columnwidth]{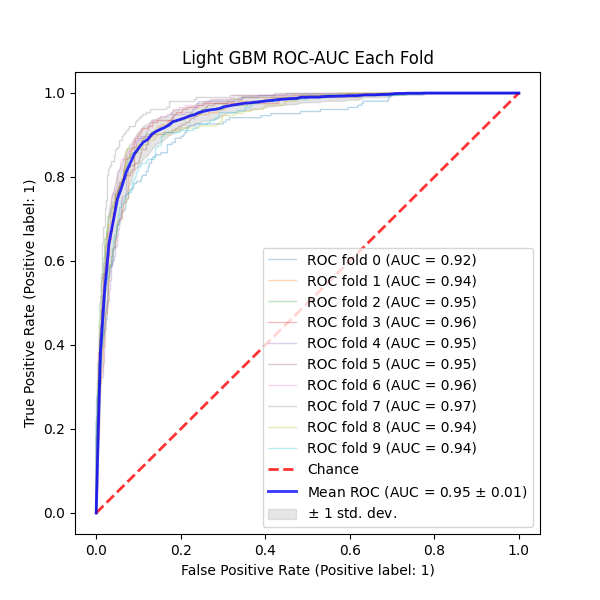}
        \label{fig:second_sub}
    }
    \caption{(a) XGBoost and (b) LightGBM results for eICU AUC-ROC}
    \label{fig:sample_subfigures}
\end{figure}
In summary, LightGBM and XGBoost models excel in both the eICU and Hospital datasets, offering a strong blend of high AUC-ROC and computational efficiency. They thus present themselves as robust choices for sepsis prediction tasks across varied datasets.

\section{Discussion}

Our models, specifically LightGBM and XGBoost, performed exceptionally well on the eICU dataset, achieving AUC-ROC and Recall scores above 0.964 and 0.827, respectively. However, performance varied when applied to different datasets, influenced by factors like data size, feature set, and class imbalance.

\subsection{St. Mary's Hospital's Dataset}
Although the model performed well on the Hospital dataset, its smaller size and label imbalance raise concerns about overfitting and generalizability, limiting its real-world applicability.

\subsection{Data Preprocessing Challenges}
We faced preprocessing challenges, such as outliers in qSOFA scores and the exclusion of the GCS predictor. The removal of outliers, while helpful in the short term, could introduce bias and affect generalizability. The omission of GCS could compromise the model's accuracy and make cross-dataset comparisons challenging.

\section{Conclusion}

In this paper, we tackled the problem of detecting sepsis in advance, with the aim to assist hospitals in effectively monitoring and managing patients' diseases. We were motivated by the need to improve the accuracy and efficiency of sepsis detection, ultimately reducing the mortality rate and healthcare burden associated with this condition.

Our methodology consisted of a comprehensive data preprocessing approach, which involved using a lag of statistical features and qSOFA score for outlier detection. We then employed two gradient boosting models, XGBoost and LightGBM, to perform the classification task. These models demonstrated promising results in the classification of sepsis onset cases, offering valuable insights for medical practitioners and researchers alike.

The main takeaway messages from our study are the effectiveness of the proposed preprocessing techniques and the potential of utilizing gradient boosting models like XGBoost and LightGBM. Moreover, the combination of the lag of statistical features. Moreover, the potential and effectiveness to applying machine learning models on different demographics.

For future research, we suggest:
\begin{enumerate}
\item Applying our methodology to other medical conditions to test its generalizability.
\item Using similar labels and a balanced dataset to tackle dataset imbalance and improve model reliability.
\item Delving deeper into deep learning models like LSTM networks for better classification. The article hasn't extensively discussed this due to differences in preprocessing and data dimensions.
\item Leveraging language models for intuitive model prediction explanations.
\end{enumerate}


In conclusion, our study has showcased the potential of utilizing machine learning techniques, particularly gradient boosting models, for the classification of sepsis onset when applied to an extended dataset encompassing various demographics.

By pursuing the suggested future directions, we hope to further enhance the performance of these models and make a meaningful contribution to improved patient care and management on a global scale.

\section*{Acknowledgement}{
We express our sincere gratitude to the St. Mary's Hospital, Seoul for generously providing the dataset for our research. Their contribution has been invaluable to the advancement of our work. We also acknowledge and respect the KIRB number (KIRB-20230120-153) associated with this data, ensuring its ethical use throughout our study.
}

\bibliographystyle{IEEEtran}
\bibliography{sample.bib}

\end{document}